\documentclass{article}

\usepackage{microtype}
\usepackage{graphicx}
\usepackage{subfigure}
\usepackage{booktabs} 

\usepackage{hyperref}

\usepackage[accepted]{icml2020}

\icmltitlerunning{E-Stitchup: Data Augmentation for Pre-Trained Embeddings}

\begin{document}

\twocolumn[
\icmltitle{E-Stitchup: Data Augmentation for Pre-Trained Embeddings}



\icmlsetsymbol{equal}{*}

\begin{icmlauthorlist}
\icmlauthor{Cameron R. Wolfe}{sfdc}
\icmlauthor{Keld T. Lundgaard}{sfdc}
\end{icmlauthorlist}

\icmlcorrespondingauthor{Cameron R. Wolfe}{wolfe.cameron@rice.edu}

\icmlaffiliation{sfdc}{Salesforce Einstein; Cambridge, MA, USA}

\icmlkeywords{Machine Learning, ICML}

\vskip 0.3in
]



\printAffiliationsAndNotice{}  

\begin{abstract}
In this work, we propose data augmentation methods for embeddings from pre-trained deep learning models that take a weighted combination of a pair of input embeddings, as inspired by Mixup \cite{mixup}, and combine such augmentation with extra label softening. These methods are shown to significantly increase classification accuracy, reduce training time, and improve confidence calibration of a downstream model that is trained with them. As a result of such improved confidence calibration, the model output can be more intuitively interpreted and used to accurately identify out-of-distribution data by applying an appropriate confidence threshold to model predictions. The identified out-of-distribution data can then be prioritized for labeling, thus focusing labeling effort on data that is more likely to boost model performance. These findings, we believe, lay a solid foundation for improving the classification performance and calibration of models that use pre-trained embeddings as input and provide several benefits that prove extremely useful in a production-level deep learning system. 
\end{abstract}


\section{Introduction}
\label{intro}

In many deep learning applications, the ability to generalize from one distribution of data to another is quite difficult. Due to the lack of alignment between training and testing distributions, many deep learning models struggle with overfitting and fail to perform well in real world or test environments, despite high training accuracy. Many approaches have been proposed to mitigate the issue of overfitting; such as leveraging transfer learning, applying data augmentation, and utilizing label softening; but the generalization performance of deep learning models is still in need of further improvement.

Transfer learning has proved an extremely useful tool for reducing overfitting in deep learning. In computer vision (CV), large, pre-trained convolutional neural networks (CNNs) have been shown, when fine-tuned to accomplish some downstream task, to perform better than CNNs trained from scratch \cite{imagenet, CNN_transfer}. Similarly, deep learning models in natural language processing (NLP) have began to heavily rely upon self-supervised pre-training approaches \cite{transformer_paper, bert_paper}. Additionally, one can use embeddings created by a pre-trained model as input to a different model that can be trained to perform some downstream task (e.g., clustering or classification) instead of fine-tuning the pre-trained model directly \cite{imgembed, bertembed}.

Data augmentation has been shown to significantly reduce overfitting in deep learning. For image data, such augmentations typically take the form of either geometric or color augmentations on input images \cite{autoaugment, generic_data_aug}. More recently, Mixup \cite{mixup}, which performs augmentation by taking a weighted average of two input images, was proposed and shown to improve model performance \cite{understanding_mixup, on_mixup}. Data augmentations have also been proposed for textual data but have been limited due to the difficulty of making the semantic meaning of a sentence or phrase robust to perturbation \cite{data_aug_machine_translation, contextual_data_aug}. In this work, the proposed forms of data augmentation are applied directly to embedding inputs instead of raw data, thus making the methodology applicable to numerous forms of data (e.g., images, text, video, speech, etc.).

Confidence calibration, defined as a model's prediction confidence matching its likelihood of correctness, allows prediction confidence to both determine whether a prediction is reliable and be used to identify out-of-distribution (OOD) data (i.e., data belonging to an unknown class), which should be assigned low confidence. Current approaches to confidence calibration typically apply parametric transformations, optimized over a validation set, to a network's penultimate layer to produce a calibrated output distribution, separate from the model's true output \cite{ODIN, nn_cal}. Using the calibrated output distribution, threshold heuristics can then be applied to detect OOD data \cite{mehlanobis_dist, ODIN, abnormality_module}. Poor calibration is typical of modern deep learning architectures \cite{nn_cal} and one-hot labels have been shown to further damage model calibration \cite{labelsmoothing_analysis}. Recent studies have shown that label softening can both improve calibration  and mitigate overconfidence (i.e., predicting all data with high confidence) \cite{labelsmoothing_analysis}. In this work, we show that our proposed forms of data augmentation, optionally combined with label softening, improve both model calibration and the detection of OOD data, without requiring any separate modules or modifications to the model itself.

The novel ideas presented in this work are as follows:
\begin{enumerate}
  \item Four variants of data augmentation are explored that utilize weighted combinations of embeddings from pre-trained networks.
  \item Embedding augmentation methods are comprehensively analyzed and shown to improve model calibration and accuracy. 
\end{enumerate}

The paper is organized in the following fashion. First, the methods are described, including several variants of data augmentation that are applicable to pre-trained embeddings. Following the methods, the experimental details and results will be outlined, forming a comparison between the proposed methodology and the control (i.e., model trained without embedding augmentation). Next, an in-depth analysis of the proposed methods is presented, followed by possible ideas for future research. Lastly, the major conclusions of the work are summarized.


\section{Methods} \label{methods}

\subsection{Creating Text and Image Embeddings} \label{emb_sec}
For both text and image data, it is possible to use pre-trained deep learning models to produce embedding vectors, which quantitatively describe data in the activation space of the pre-trained network. For textual data, recently proposed transformer architectures, such as BERT \cite{bert_paper, bertembed} and XLNet \cite{xlnet}, have been shown to be effective at producing embeddings. Modern CNN architectures such as ResNets \cite{resnet} or EfficientNet \cite{EfficientNet} can be used to produce useful image embeddings. Such embeddings are created by passing the image or textual data as input to a pre-trained model and using its hidden layer activations as embeddings. Because the final layers of such pre-trained models generally contain high-level, semantic information about the input data, embedding vectors are typically taken from these final layers. However, including lower-level information from earlier activation layers may sometimes be useful.

\subsection{E-Mixup} \label{emb_mix}
\begin{figure}
\includegraphics[width=3.0in]{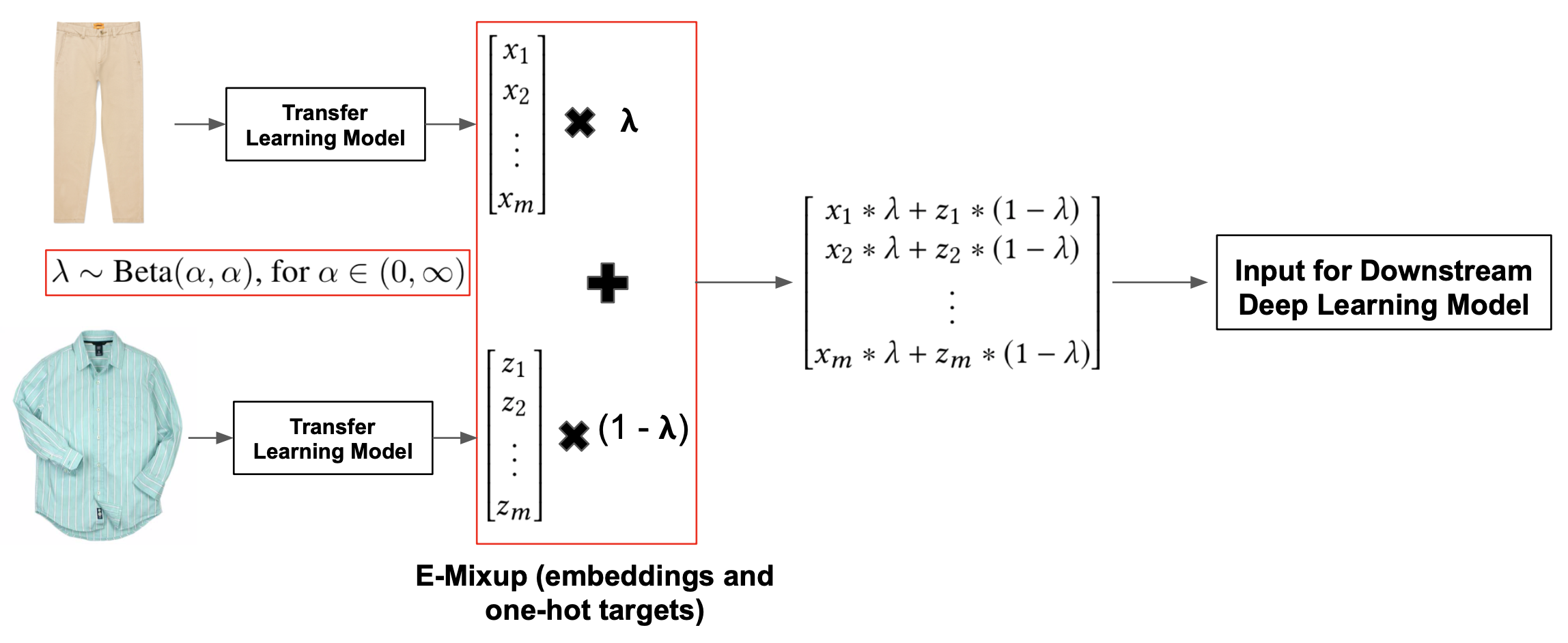}
\caption{Outlines how embedding representations are augmented using E-Mixup. Both the embedding inputs and their associated target vectors (i.e., one-hot prediction targets) are augmented using the same process outlined above.}
\label{embmix_fig}
\end{figure}
The first form of data augmentation proposed in this work is referred to as E-Mixup, which stands for "Embedding Mixup". The idea of E-Mixup is inspired by Mixup data augmentation \cite{mixup}, but operates on embeddings instead of raw images. In E-Mixup, a random value, lambda, is first sampled from a Beta distribution having a parameter alpha (i.e., alpha refers to this distribution parameter, not the learning rate). Once lambda has been sampled, E-Mixup takes a weighted average over the embedding inputs of two unique training examples, where lambda is the weight of the average. Each time two inputs are combined, a new lambda value is randomly sampled. The same process is applied to both input embeddings and their associated target vectors, thus replacing one-hot targets with the weighted average of the two labels. This process can be visualized in Fig. \ref{embmix_fig}.

\subsection{E-Stitchup} \label{stitchup}
\begin{figure}
\includegraphics[width=3.0in]{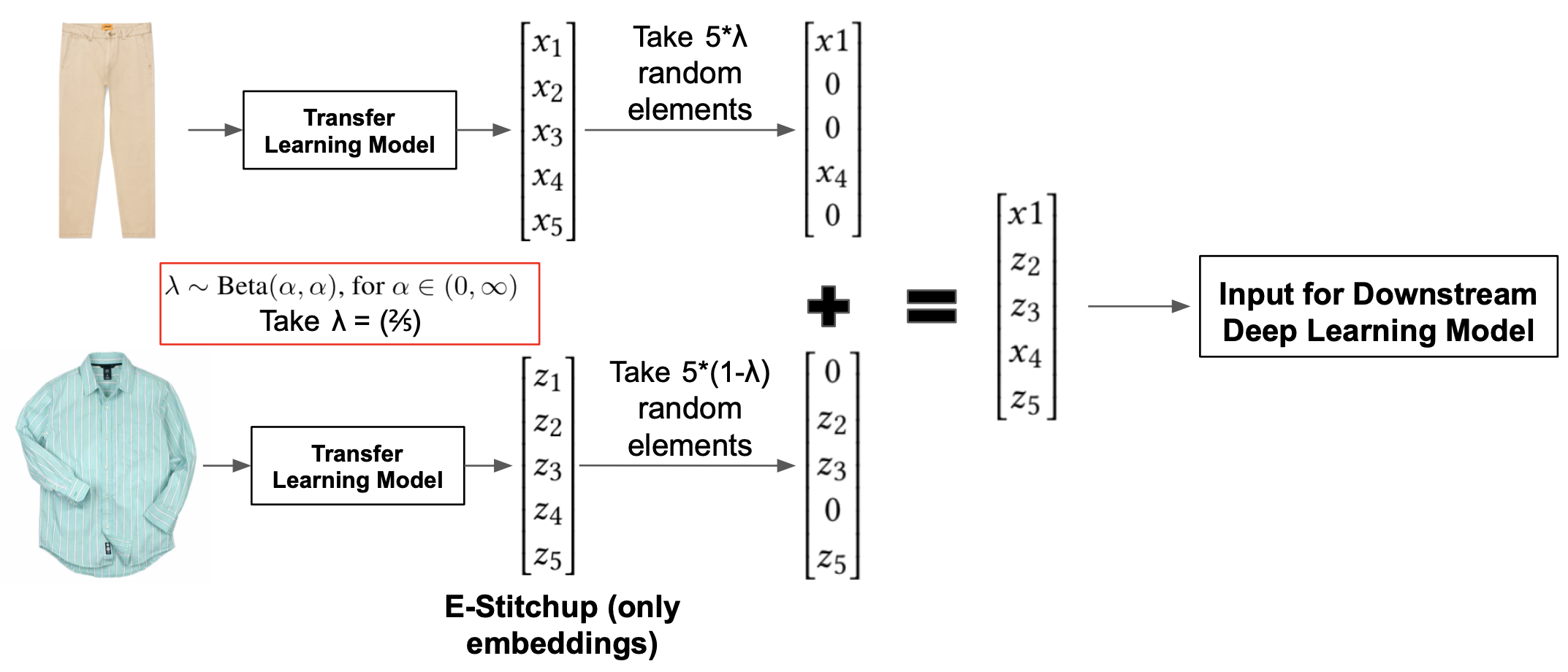}
\caption{Outlines how embedding representations are augmented in E-Stitchup. The label vectors associated with these embedding inputs are not handled in the same way, but are instead handled by taking a weighted average of the two label vectors, as in E-Mixup.}
\label{stitchup_fig}
\end{figure}
The second proposed form of data augmentation is referred to as E-Stitchup. E-Stitchup was inspired by recent experiments that perform Mixup by randomly pasting a cropped section of one image over another instead of taking a pixelwise average \cite{understanding_mixup}. E-Stitchup combines two embeddings by randomly selecting the value at each index of the combined embedding from the value at that index for one of the two original embeddings. The probability of choosing an element from either vector is determined by a value, lambda, that is sampled from a random beta distribution with parameter alpha. The target vector is created by taking a weighted average of the two original target vectors, as in E-Mixup. The process of E-Stitchup can be observed in Fig. \ref{stitchup_fig}. 

\subsection{Softened Embedding Augmentation} \label{soft_sec}

Two new variants of embedding augmentation are created by combining the proposed methods with extra label softening. Although embedding augmentation already eliminates one-hot labels by taking a weighted average of target vectors, label softening augments this effect by further softening the resulting target vector. In this work, label softening is performed by first subtracting a small value from classes with nonzero probability. This subtracted value is referred to as the "Label Softness" and the subtraction is clamped at zero to never yield a negative probability. Additionally, a total probability of one is distributed across all classes with zero probability. The sum of probabilities in the target distribution is not constrained to a value of one due to the use of a binomial output transformation instead of softmax in the classification model (See Sec. \ref{model_sec}). The augmentation methods with added label softening are referred to as Soft E-Mixup and Soft E-Stitchup, respectively.

\subsection{The "None" Category} \label{none_sec}
The proposed forms of embedding augmentation are partially evaluated on the ability of resulting models to detect OOD data. In this work, OOD data is identified by applying a threshold to network outputs, referred to as the "Confidence Threshold", such that a prediction is only valid if the model assigns probability greater than this threshold to a certain class. If multiple classes are given probabilities greater than the confidence threshold, the class with the greatest probability is selected. If no classes are assigned probabilities greater than the confidence threshold, the data is considered OOD, or part of the "none" category. It should be noted that, throughout this work, all categories that are not within the "none" category are referred to as in-distribution (ID) categories, while OOD is synonymous with the "none" category. The accurate identification of "none" category data is a major goal of this work and is closely related to the problem of confidence calibration. Unlike recent approaches to confidence calibration and identifying OOD data \cite{nn_cal, mehlanobis_dist, abnormality_module, ODIN}, our work directly calibrates network outputs by using embedding augmentation to incorporate confidence calibration into the model's training procedure.


\section{Experimental Details} \label{exp_details}
Several experiments are performed to analyze the effectiveness of the proposed embedding augmentation methods. Each experiment is repeated for five trials with different training and validation splits to ensure the consistency of the results. The best performing hyperparameters for each experiment (i.e., alpha value and label softness) are determined by first measuring performance on a small validation set that is taken randomly from the training data for a given trial.  In future revisions to the work, hyperparameters will instead be selected with hierarchical bootstrap, a cross validation technique that is more thorough than a randomly-chosen validation set \cite{mBEEF}. The control experiment, presented alongside the proposed augmentation methods, corresponds to an experiment in which no embedding augmentation is used, while the soft control experiments corresponds to an experiment with label softening but no embedding augmentation.

\subsection{Fashion Product Images Dataset} \label{dataset}
The Fashion Product Images dataset, which is available at \cite{FashionData}, is used for all experiments in this work. This dataset contains data for 44K apparel products, each of which has an associated image, product title, and product description. These products are classified into 171 unique categories. The textual data and images are converted into their associated embedding representations before being fed as input to the downstream classification model (see Sec. \ref{emb_det}). This dataset was selected so that the proposed methodology could be simultaneously applied to both text and image embeddings. Additionally, the products in this dataset are quite useful because they closely resemble items that would be classified in an industrial application, such as an e-commerce recommendation system.

\subsection{Pre-Trained Embedding Models} \label{emb_det}
All textual embeddings were created with the BERT Base transformer model (i.e., HuggingFace PyTorch implementation) \cite{bert_paper}. This model was never fine-tuned or modified in any way. To create the embeddings, input phrases are tokenized using a WordPiece tokenizer \cite{WordPiece}, converted into token embeddings, and fed as input to BERT. A textual embedding is then created by finding the average output vector of the last and second-to-last layer of the transformer and concatenating these average vectors together, resulting in an embedding of size 1536. If there are multiple phrases or descriptions associated with a single data element (e.g., both a product title and description), embeddings are created separately for each of these phrases and then concatenated.

Image embeddings were created with the EfficientNet B4 model \cite{EfficientNet}, which was never fine-tuned or modified in any way. After passing an image as input to the CNN, global average pooling is applied to the CNN's final convolutional layer to create an image embedding of size 1792. In cases where both image and textual data are available, all image and textual embeddings are created separately and concatenated together before being passed as input to the downstream model.

\subsection{Classification Model} \label{model_sec}
The downstream classification model is a fully-connected network that accepts a fixed-size input and outputs a probability distribution over all possible classes. This model has two hidden layers of size 250 that are both followed by a Dropout layer \cite{dropout} with probability of 0.3 and a Rectified Linear Unit activation layer. A binomial output transformation (i.e., element-wise sigmoid operation) is applied to the model's output layer before the predicted class is determined. This binomial output transformation is used instead of softmax because it enables the model to assign low probability to all classes, thus simplifying the handling of OOD data. This downstream model is significantly smaller than most pre-trained models used for transfer learning and can be retrained at a low computational cost.

\subsection{Training Details}
For every experiment, weight decay is set to .0001 and a linear learning rate cycle is utilized that fluctuates between a learning rate of 0.0003 and 0.003 every 12 epochs. Training is continued for 576 epochs for all experiments to ensure convergence. Only 10\% of the available data is used for training, while the rest of the data is used for validation. A small training set is used to simulate a scenario with limited training data, which increases the likelihood of overfitting. Because the training set is so small, entire classes of products may be excluded from the training set, which allows "none" category accuracy to be evaluated as described in Sec. \ref{none_sec}. Each experiment is performed with a different training and validation split to ensure the consistency of the results. 

\subsection{Accuracy Metric} \label{acc_sec}
There exist multiple manners in which the accuracy metric can be defined in this work. For all results presented, accuracy is considered to be the top-one accuracy of class probabilities (i.e., the highest-probability class in the model's output layer is the predicted class). However, if the prediction confidence is lower than the confidence threshold, the product is considered to be in the "none" category. Similarly, all product classes that are not present in the training set, but are present in the validation set, are considered part of the "none" category. A correct classification occurs when either the top-one prediction, with probability above the confidence threshold, is equal to a product's labeled class, or no classes are assigned a probability higher than the confidence threshold for a product in the "none" category.


\section{Results} \label{result_sec}
\begin{figure*}[!ht]
\includegraphics[width=6.5in]{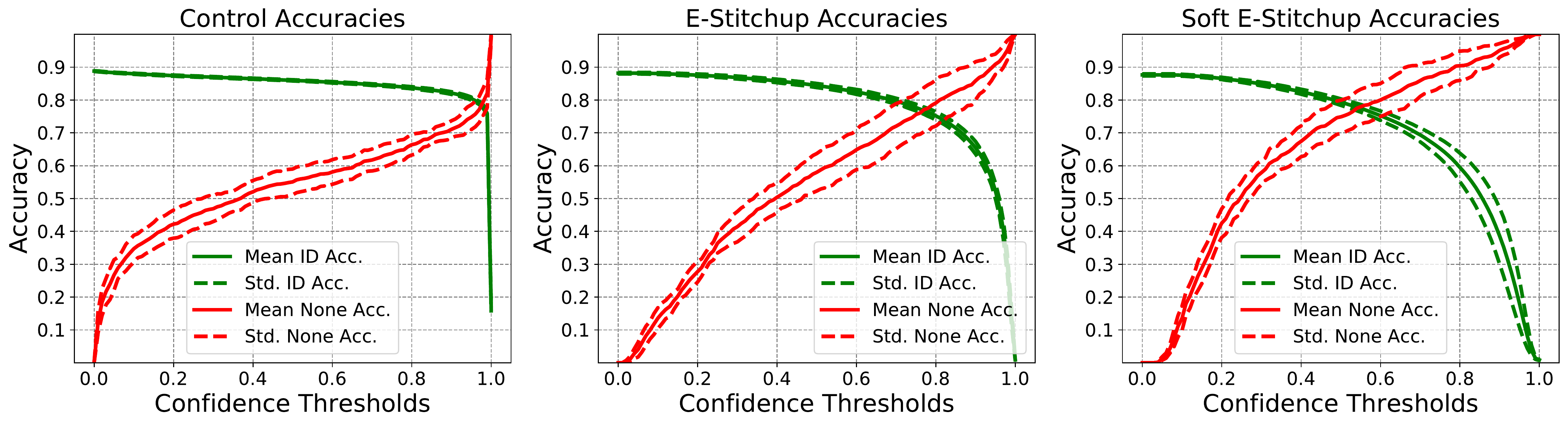}
\caption{In-distribution (i.e., all products that are not in the "none" category) and "none" category accuracy across different confidence thresholds for the control experiment (top), E-Stitchup (middle), and Soft E-Stitchup (bottom). The solid lines represent average accuracy across all trials, while the dotted lines represent the standard deviation in accuracy between trials.}
\label{acc_metrics}
\end{figure*}

\begin{figure*}[!ht]
\includegraphics[width=6.5in]{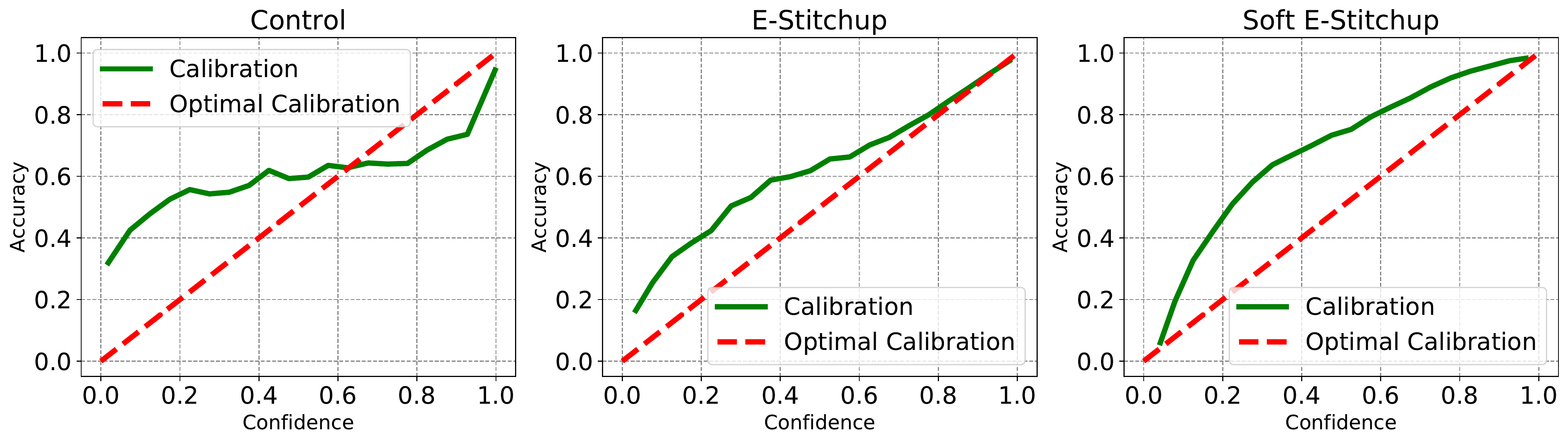}
\caption{Reliability diagrams for models trained in the control experiment (top), E-Stitchup experiment (middle), and Soft E-Stitchup experiment. The accuracy and confidence metrics presented in these reliability diagrams are averaged across all trials of each experiment.}
\label{reliability_fig}
\end{figure*}

\begin{table}[ht]
\begin{center}
\begin{tabular}{r c c}
Method & AUROC & AUPR \\
 \hline
 Control & \phantom{+}0.9698 (0.0010) & \phantom{+}0.8913 (0.0026)\\ 
 Soft Control & +0.0200 (0.0006) & +0.0196 (0.0028) \\
 E-Mixup & +0.0173 (0.0014) & +0.0164 (0.0054) \\
 E-Stitchup & +0.0189 (0.0007) & +0.0222 (0.0031)  \\
 Soft E-Mixup & +0.0216 (0.0008) & +0.0228 (0.0022)  \\
 Soft E-Stitchup & \textbf{+0.0226 (0.0012)} & \textbf{+0.0236 (0.0020)} \\
\end{tabular}
\end{center}
 \caption{Weighted average AUROC and AUPR scores across all product categories, including the "none" category, for models trained with each augmentation method. The values presented are all relative to the average AUROC and AUPR score of the control experiment. The deviations listed represent the standard deviation of a method's relative improvement over the control across all trials.}
  \label{auc_table}
\end{table}

\begin{table}[ht]
\begin{center}
\begin{tabular}{r c} 
Method & Conf.-Acc. Correlation \\
 \hline
 Control & 0.9072 \\
 Soft Control & 0.9292 \\
 E-Mixup & 0.9789 \\
 E-Stitchup & \textbf{0.9860} \\
 Soft E-Mixup & 0.9236  \\
 Soft E-Stitchup & 0.9472 \\
\end{tabular}
\end{center}
 \caption{Confidence-accuracy correlation for models trained with different variants of embedding augmentation, as well as for the control experiment. This metric reflects the degree of calibration in a model's predictions.}
  \label{calib_table}
\end{table}

In this section, each of the proposed augmentation methods are evaluated in comparison to the control. Although accuracy metrics are presented (See Fig. \ref{acc_metrics}), these values are dependent upon the value of the confidence threshold (see Sec. \ref{acc_sec}). Therefore, the Area Under the Receiver Operating Characteristic Curve (AUROC) and Area Under the Precision-Recall Curve (AUPR), which are threshold-independent metrics, are mostly used to evaluate model performance. Additionally, confidence calibration is assessed by plotting reliability diagrams \cite{nn_cal} and measuring the confidence-accuracy correlation for the predictions of each model.

\subsection{E-Mixup and Soft E-Mixup}
\begin{figure}
\includegraphics[width=3.0in]{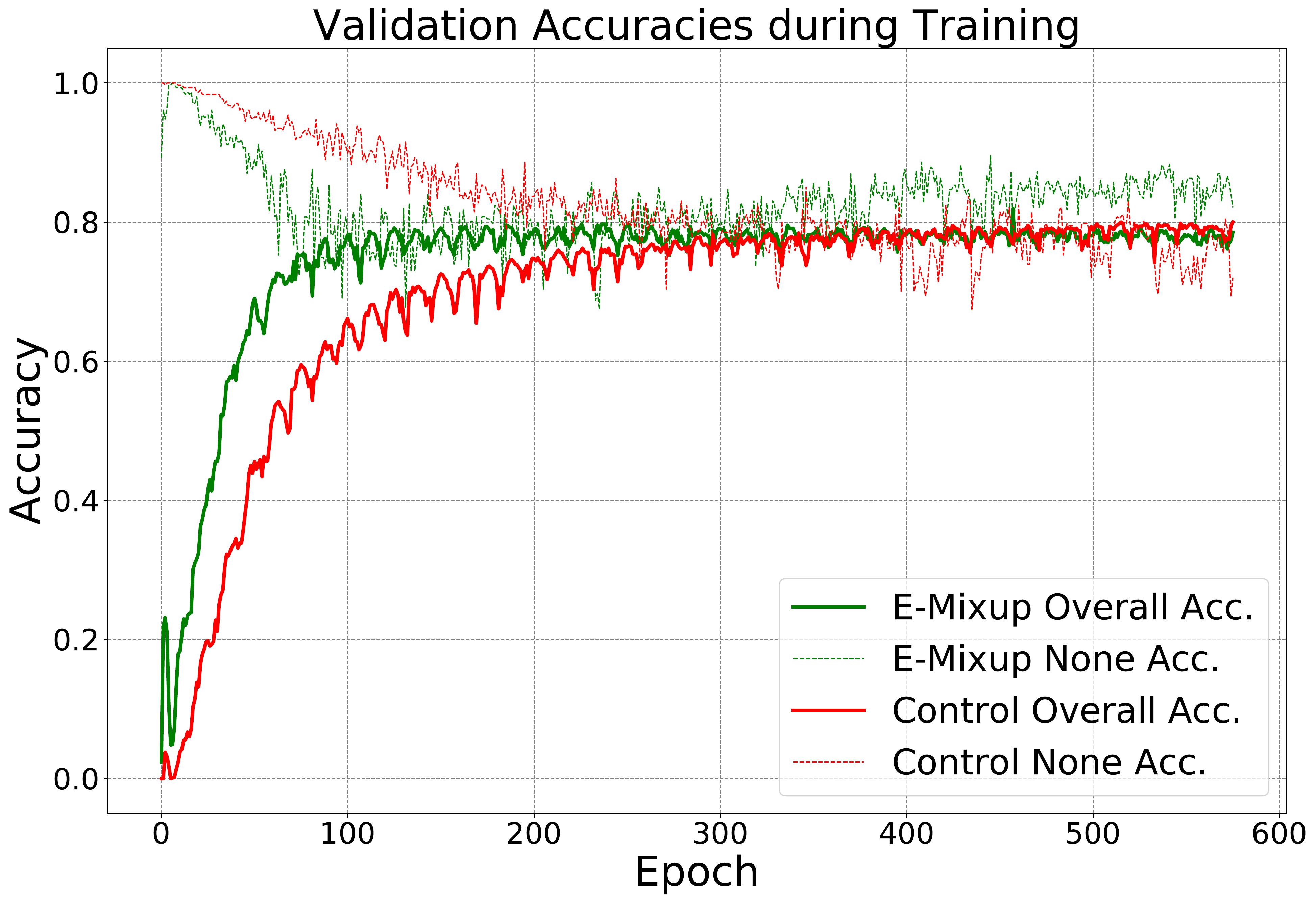}
\caption{Overall and "none" category accuracy throughout training for control model and E-Mixup model.}
\label{long_train}
\end{figure}
As seen in Table \ref{auc_table}, E-Mixup has an improvement of 0.0173 and 0.0164 over the control experiment's AUROC and AUPR, respectively. However, E-Mixup yields the lowest AUROC and AUPR of all augmentation experiments. If extra label softening is added to E-Mixup (i.e., Soft E-Mixup), the resulting model improves upon the control experiment's AUROC and AUPR by 0.0216 and 0.0228, respectively. Additionally, models trained with both E-Mixup and Soft E-Mixup have improved calibration relative to the control. To quantitatively demonstrate the improved calibration of models trained with embedding augmentation, the correlation between model accuracy and confidence is measured, as seen in Table \ref{calib_table}. Models trained with E-Mixup and Soft E-Mixup yield confidence-accuracy correlations of 0.9789 and 0.9236, while the control experiment yields a correlation of 0.9072.

As can be seen in Fig. \ref{long_train}, models trained with E-Mixup converge in half the number of epochs compared to the control and maintain a stable "none" category accuracy throughout later epochs in training, allowing accurate and consistent classification of OOD data.  Faster convergence and stable "none" category accuracy, although demonstrated with E-Mixup in Fig. \ref{long_train}, were found to be common characteristics of models trained with all forms of embedding augmentation. Models trained without embedding augmentation (i.e., both control and soft control experiments) tend to have volatile "none" category accuracy that deteriorates throughout training and take significantly longer to converge.

\subsection{E-Stitchup and Soft E-Stitchup}
E-Stitchup yields an improvement of 0.0189 and 0.0222 in AUROC and AUPR relative to the control, while Soft E-Stitchup yields an improvement of 0.0226 and 0.0236 in AUROC and AUPR, which is the highest recorded performance of any experiment. E-Stitchup and Soft E-Stitchup also yield improvements in confidence calibration. As seen in Table \ref{calib_table}, models trained with E-Stitchup achieve a confidence-accuracy correlation of 0.9860, which is the highest recorded correlation of any experiment, while models trained with Soft E-Stitchup yield a correlation of 0.9472. The improved calibration of models trained with E-Stitchup is illustrated in the associated reliability diagram in Fig. \ref{reliability_fig}.

E-Stitchup yields higher performance than E-Mixup both in classification performance and model calibration. Such a discovery is interesting because it suggests that performing Mixup on embedding data requires adaptations to the normal method of Mixup \cite{mixup} to achieve optimal performance. In this case, it becomes clear that sampling embedding values yields better performance than interpolating between them.

\subsection{Setting the Confidence Threshold}
\begin{figure}
\includegraphics[width=3.0in]{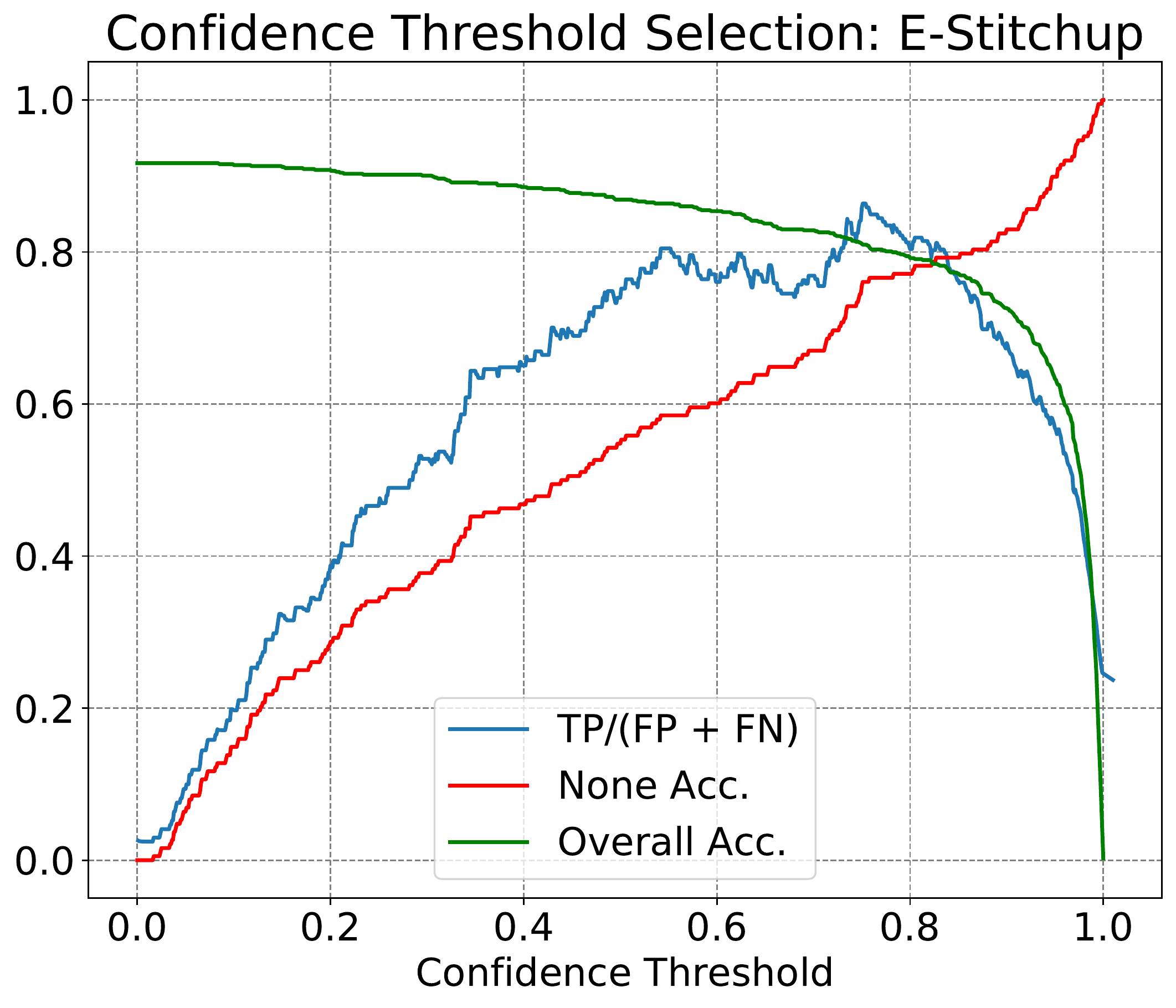}
\caption{Displays the average ratio of true positives to false negatives and false positives between ID categories and the "none" category, as well as accuracy across confidence thresholds. These plots display two heuristics for choosing a useful confidence threshold.}

\label{select_ct}
\end{figure}
\begin{table}[ht]
\begin{center}
\begin{tabular}{r c c c c}
 Method & Thresh. & ID Acc. & None Acc. & Acc. \\
 \hline
 Control & 0.93 & 0.82 & 0.72 & 0.82\\
 S. Control & 0.50 & 0.80 & 0.75 & 0.80 \\
 E-Mixup & 0.70 & 0.79 & 0.75 & 0.79 \\
 E-Stitchup & 0.78 & 0.78 & 0.77 & 0.78\\
 S. E-Mixup & 0.52 & 0.75 & 0.80 & 0.75\\
 S. E-Stitchup & 0.58 & 0.76 & 0.79 & 0.76\\
\end{tabular}
\end{center}
 \caption{The ID, "none" category, and overall accuracy of models trained with each augmentation method, as well as the control models, after the optimal confidence threshold has been selected. Overall accuracy is roughly equal to ID accuracy because the number of products in the "none" category is much smaller than the number of ID products.}
  \label{opt_perf}
\end{table}

To obtain optimal performance with the proposed methodology, the correct value for the confidence threshold must be chosen. As seen in Fig. \ref{select_ct}, two heuristics are developed to estimate an optimal confidence threshold. Both heuristics require the fabrication of a small validation set that contains OOD data. In this work, the validation set was constructed by creating a 70-30 split for training and validation data and choosing five random product categories to be fully removed from the training set, thus ensuring the existence of "none" category products in the validation set. The model's predictions on this validation set can be used to estimate the optimal confidence threshold. 

The first heuristic finds the intersection between overall and "none" category accuracy, represented by the green and red curves in Fig. \ref{select_ct}, which, in the provided example, occurs at a threshold of 0.8. The second heuristic plots the ratio between true positive predictions and the sum of false positive and false negative predictions. This ratio is computed separately for the "none" category and ID categories so that the "none" category can be equally weighted. The optimal confidence threshold can then be determined by finding the maximum value of this ratio, which, in Fig. \ref{select_ct}, occurs at 0.77. Although these heuristics yield slightly different confidence threshold estimates, their accuracy is generally robust to the size of the validation set, which allows an optimal confidence threshold to be estimated consistently.

The performance of each method after its optimal confidence threshold is chosen using the heuristics defined above can be seen in Table \ref{opt_perf}. At the optimal confidence threshold, E-Stitchup seems to achieve the best balance between overall and "none" category accuracy. In certain cases, however, methods that seem to perform higher in "none" accuracy, such as Soft E-Mixup, or ID accuracy, such as the control, may be preferred.


\section{Analysis}
\subsection{Classification Performance} \label{classif_sec}
From the provided results in Table \ref{auc_table}, it is clear that embedding augmentation results in improved classification performance. E-Stitchup is shown to produce models with the best performance, but adding extra label softening (i.e., Soft E-Stitchup) consistently improves classification performance in terms of AUROC and AUPR. The accuracy metrics provided for each of the experiments in Table \ref{opt_perf} show that the control experiments are surprisingly competitive with embedding augmentation methods. However, the AUROC and AUPR measures, which give an unbiased view of classification performance, show a slight, but clear improvement in classification performance using embedding augmentation, especially when extra label softening is added. Additionally, embedding augmentation has the added benefit of reducing the number of epochs before convergence, as seen in Fig. \ref{long_train}.

Models trained with embedding augmentation achieve a smooth tradeoff in "none" category and ID accuracy across different confidence thresholds (see Fig. \ref{acc_metrics}). In the control, models favor overall accuracy and yield low "none" category accuracy across nearly all confidence thresholds. However, model performance changes rapidly at a threshold of ~0.98 because the majority of the control model's predictions are of high confidence. Such rapid changes in accuracy across minimal changes in confidence threshold highlight the volatility of the control model's performance. Additionally, because the control favors ID accuracy over "none" category accuracy at most confidence thresholds, it is nearly impossible to choose a confidence threshold for the control that prioritizes "none" category accuracy. No such issues arise in models trained with embedding augmentation, all of which yield classification behavior resembling Fig. \ref{acc_metrics} for normal and softened embedding augmentation methods, respectively. Models trained with embedding augmentation can find a stable balance between performance on ID products and "none" category products and are capable of sacrificing ID accuracy to achieve higher levels of "none" category accuracy. Such a wider scope of performance possibilities, which cannot be achieved in the control, can prove useful in accurately identifying OOD data or data that a model does not understand well. 

In analyzing the classification behavior of the proposed embedding augmentation variants, a few observation arise that provide useful insight into their performance. Interestingly, models trained with softened embedding augmentation achieve an intersection of ID and "none" category accuracy at a lower confidence threshold compared to those trained with normal embedding augmentation (see Fig. \ref{acc_metrics}). This intersection at a decreased confidence threshold, which allows an even wider scope of "none" and ID accuracies to be obtained, demonstrates the impact of label softening in regularizing the confidence of model predictions. Furthermore, the improved performance of E-Stitchup in comparison to E-Mixup is quite interesting. Such improved performance suggests that deep learning models trained with embedding input perform better when the values within these embeddings are preserved instead of interpolated. This improvement in performance was not as noticeable when applied to raw pixel values of images in previous studies \cite{understanding_mixup}. Therefore, these observations highlight that the original Mixup methodology \cite{mixup}, which was developed for raw image inputs, must be slightly adapted for optimal performance on embedding data. Such improved methodology should continue to be studied and developed because it allows the idea of Mixup to be applied to numerous kinds of data, such as images, text, speech, or any other kind of data for which an embedding can be created. 

\subsection{Confidence Calibration} \label{cc_sec}
\begin{figure}
\includegraphics[width=3.0in]{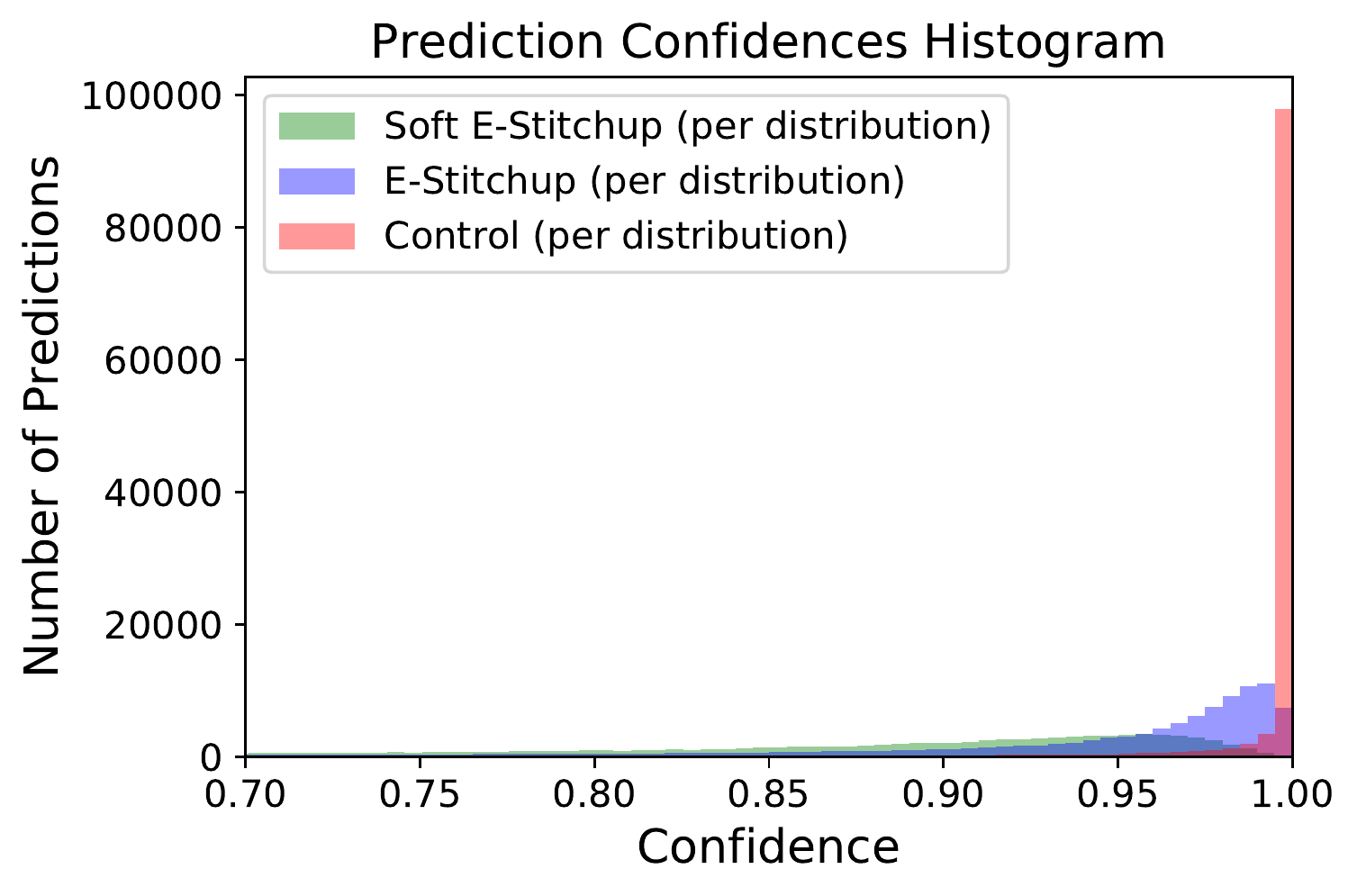}
\caption{A histogram of prediction confidences for the control experiment, E-Stitchup, and Soft E-Stitchup.}
\label{conf_hist}
\end{figure}
The model trained without any embedding augmentation shows symptoms of severe overconfidence that damage its performance. As seen in Fig. \ref{conf_hist}, almost all predictions made by the control model are of high confidence. Such overconfidence leads to the volatile performance of the control model observed in Fig. \ref{acc_metrics}, causing ID and "none" category accuracy to shift rapidly at a confidence threshold of ~0.98. Overconfidence is common for models trained with one-hot labels because small errors in confidence (i.e., a prediction of confidence 0.95 instead of 1.0) can dominate the gradient during training and lead a model to produce peaked output distributions. As seen in Fig. \ref{conf_hist}, embedding augmentation, which eliminates one-hot labels (see Sec. \ref{methods}), clearly regularizes overconfidence, causing model predictions to no longer be clustered at high confidence. Additionally, label softening, when combined with embedding augmentation, leads to an even wider distribution of prediction confidences, thus eliminating issues with overconfidence that are seen in the control model.

Models trained with embedding augmentation have improved confidence calibration (see Table \ref{calib_table}). As seen in Fig. \ref{reliability_fig}, the accuracy in the reliability diagram of the control model remains nearly constant across all confidence levels and peaks at high confidence, again highlighting the control model's issues with overconfidence. In contrast, models trained with embedding augmentation demonstrate a clear, direct relationship between accuracy and confidence, achieving calibration that is much closer to optimal. This direct relationship is confirmed by the confidence-accuracy correlation of models trained with embedding augmentation, all of which exceed the correlation values of control experiments. Interestingly, adding label softening to embedding augmentation results in decreased model calibration, despite improving classification performance. It can be seen in Fig. \ref{reliability_fig} that such decreased calibration relative to other augmentation methods is caused by model underconfidence (i.e., accuracy of model predictions exceeds their associated confidence), thus further highlighting the effect of label softening on regularizing model confidence. The increased calibration of models trained with embedding augmentation is useful, as it allows network confidence values to be intuitively interpreted by a human and can lead to improved OOD data detection. 

To achieve optimal model calibration, E-Stitchup should be used during training, as it yields the best correlation between model confidence and accuracy (see Table \ref{calib_table}). As previously mentioned, however, classification performance can be further optimized at the cost of lower confidence calibration by using Soft E-Stitchup or even the control. The control experiments, both with and without label softening, achieve relatively competitive accuracy in comparison to embedding augmentation experiments, but struggle with severe overconfidence and poor confidence calibration. Therefore, the significant improvements in model calibration are shown to be a clear result of embedding augmentation and can be maximized with the use of E-Stitchup. 


\section{Discussion and Future Work}
The proposed augmentation methods are shown to improve classification performance, boost model calibration, improve OOD data identification, and even reduce training time. To understand why these benefits are useful in a production-level deep learning system, the following scenario can be considered. After training an initial model with embedding augmentation, one can easily identify OOD and low-confidence data so that labeling efforts are focused on data that, if labeled and included in the model's training set, has a high likelihood of improving model performance. As outlined above, models trained with embedding augmentation also converge quickly, allowing them to be retrained at a low computational cost. Therefore, with the use of embedding augmentation, one can dynamically identify and provide labels for data on which the model performs poorly, include such data in the model's training set, and retrain the model. This iterative process leads to relatively quick improvements in model performance on new and unknown data, which makes it quite useful in active learning scenarios (i.e., when new classes of data are being continually introduced) that are common in production-level deep learning settings. 

There exist several avenues that could be explored to expand upon this work. First, embedding augmentation can be tested with other machine learning models, such as random forests \cite{random_forest} or support vector machine \cite{svm} models. Such experiments could be used as a benchmark to the deep learning approach in this work, as well as determine the effectiveness of embedding augmentation in other machine learning domains. Additionally, further experiments could be performed with different pre-trained models, such as XLNet \cite{xlnet} or other Sesame Street models. Lastly, because the proposed methodology is highly applicable to active learning scenarios, more work can be done to explore optimal procedures for adding new classes of data into a model's training set. For example, if a new class of data is extremely similar to an existing class (e.g., pairs of pants that do or do not require a belt), this class must be identified as part of the "none" category and added, as a new class, into the model's training set without diminishing performance on the other, similar class. Such a task is still quite difficult in current deep learning systems. 

\section{Conclusions}
In this work, several variants of data augmentation are developed for pre-trained embeddings. The proposed methods are generally applicable to embedding representations of data, making them useful in numerous domains (e.g., CV, NLP, speech processing, etc.). E-Stitchup is the most useful embedding augmentation technique. By using the proposed heuristics to select a confidence threshold, E-Stitchup can simultaneously achieve high in-distribution and "none" category accuracy. Models trained without embedding augmentation (i.e., the standard approach) struggle to achieve such a stable balance between these metrics because they suffer from poor calibration and overconfidence, causing them to favor in-distribution accuracy over "none" category accuracy. By using E-Stitchup during training, the resulting model will have improved classification performance, will no longer suffer from overconfidence, will have highly-calibrated output, and will converge in half the number of epochs in comparison to a model trained without E-Stitchup. We believe these augmentation methods for pre-trained embeddings, which require no fine-tuning or modification of pre-trained architectures, provide a useful framework for improving the performance of downstream deep learning models that use pre-trained embeddings as input, while adding minimal extra cost into the training process.

\bibliography{citation}
\bibliographystyle{icml2020}

\end{document}